\documentclass[10pt,twocolumn,letterpaper]{article}

\usepackage{cvpr}              % To produce the CAMERA-READY version
\newcommand{\paragrapht}[1]{\noindent\textbf{#1}}

\usepackage{comment}
\usepackage{lipsum}

\usepackage{amssymb}
\usepackage{xcolor}

\newcommand{\greencheck}{{\color{ForestGreen}\ding{51}}}
\newcommand{\blackcross}{\color{Maroon}{\ding{55}}} 

\usepackage{pifont}

\usepackage{algorithm}
\usepackage{algpseudocode}
\usepackage{amsmath}

\definecolor{cvprblue}{rgb}{0.21,0.49,0.74}
\usepackage[pagebackref,breaklinks,colorlinks,allcolors=cvprblue]{hyperref}

\def\paperID{*****} % *** Enter the Paper ID here
\def\confName{CVPR}
\def\confYear{2025}

\title{DBMovi-GS: Dynamic View Synthesis from Blurry Monocular Video \\ via Sparse-Controlled Gaussian Splatting}

\author{Yeon-Ji Song$^{1,3}$, Jaein Kim$^{1,3}$, Byung-Ju Kim$^{2,3}$, Byoung-Tak Zhang$^{1,2,3}$\thanks{Correspondence to btzhang@snu.ac.kr.}\\
$^1$Interdisciplinary Program in Neuroscience \quad $^2$IPAI \quad $^3$AIIS, Seoul National University \\
}

\begin{document}
\maketitle
\begin{abstract}
Novel view synthesis is a task of generating scenes from unseen perspectives; however, synthesizing dynamic scenes from blurry monocular videos remains an unresolved challenge that has yet to be effectively addressed.
Existing novel view synthesis methods are often constrained by their reliance on high-resolution images or strong assumptions about static geometry and rigid scene priors.
% , assuming limited object or camera motion.
Consequently, their approaches lack robustness in real-world environments with dynamic object and camera motion, leading to instability and degraded visual fidelity.
To address this, we propose Motion-aware \textbf{D}ynamic View Synthesis from \textbf{B}lurry \textbf{Mo}nocular \textbf{Vi}deo via Sparse\textbf{-}Controlled \textbf{G}aussian \textbf{S}platting (DBMovi-GS), a method designed for dynamic view synthesis from blurry monocular videos.
Our model generates dense 3D Gaussians, restoring sharpness from blurry videos and reconstructing detailed 3D geometry of the scene affected by dynamic motion variations.
Our model achieves robust performance in novel view synthesis under dynamic blurry scenes and sets a new benchmark in realistic novel view synthesis for blurry monocular video inputs.
\end{abstract}

% DBMovi-GS enables efficient and stable view synthesis by .
% We explicitly decompose dynamic scenes by separating object motion and camera-induced blur, preserving scene consistency across views.

\iffalse
Novel view synthesis is a task of generating scenes from unseen perspectives; however, synthesizing dynamic scenes from blurry monocular videos remains an unresolved challenge that has yet to be effectively addressed. Existing novel view synthesis methods are often constrained by their reliance on high-resolution images or strong assumptions about static geometry and rigid scene priors. Consequently, their approaches lack robustness in real-world environments with dynamic object and camera motion, leading to instability and degraded visual fidelity. To address this, we propose Motion-aware Dynamic View Synthesis from Blurry Monocular Video via Sparse-Controlled Gaussian Splatting (DBMovi-GS), a method designed for dynamic view synthesis from blurry monocular videos. Our model generates dense 3D Gaussians, restoring sharpness from blurry videos and reconstructing detailed 3D geometry of the scene affected by dynamic motion variations. Our model achieves robust performance in novel view synthesis under dynamic blurry scenes and sets a new benchmark in realistic novel view synthesis for blurry monocular video inputs.
\fi    
\vspace{-1.3em}
\section{Introduction} \label{sec:intro}
3D Gaussian Splatting (3DGS)~\citep{gaussian-splatting} has emerged as a promising tool for photorealistic scene reconstruction and novel view synthesis, offering real-time rendering, improved efficiency, and reduced training time. However, 3DGS faces challenges in complex real-world settings, particularly in dynamic, multi-object, and blurry scenes~\citep{bian2023nope,gsonthemove}.
A critical prerequisite for effective 3DGS is the accurate point cloud initialization, usually handled by the Structure-from-Motion (SfM)~\citep{sfm} implementations such as COLMAP~\citep{colmap}. Inaccuracies in SfM initialization and difficulties in simultaneous camera and object motion estimation limit its applicability to static, high-resolution inputs.
Consequently, existing 3DGS-based methods are challenged by long-exposure inputs, varying lighting conditions, and camera-induced motion blur.
Common sources of real-world blur further degrade reconstruction quality and introduce artifacts in synthesized views.
 % such as defocus blur and motion blur, 
 \begin{table}[t]
    \centering
    \setlength{\tabcolsep}{0.5mm}
    \resizebox{1\linewidth}{!}{%
    \begin{tabular}{lcccc}
    \toprule[1.5pt] 
    \multicolumn{1}{c}{} &
    \multicolumn{3}{c}{\textbf{Image Deblurring}} &
    \multicolumn{1}{c}{\textbf{View Synthesis}} \\
    \cmidrule(rl){2-4} \cmidrule(rl){5-5}
          & Defocus & Camera Mot- & Object Mot- & Camera Pose \\
    \textbf{Model} & Blur    & ion Blur    & ion Blur    & Trajectory \\
    \midrule[1pt]
    % 3DGS+Restormer\citep{Zamir2021Restormer} & \blackcross & \blackcross & \blackcross & \blackcross \\
    % Deblur-NeRF\citep{deblurnerf}            & \greencheck & \blackcross & \blackcross & \blackcross \\
    Deblur-GS\citep{chen2024deblurgs}        & \greencheck & \greencheck & \blackcross & \blackcross\\
    Deblurring 3DGS\citep{lee2024deblurring} & \greencheck & \blackcross & \greencheck & \blackcross\\
    Nope NeRF\citep{bian2023nope}            & \blackcross & \greencheck & \blackcross & \greencheck\\
    GS on the Move\citep{gsonthemove}        & \blackcross & \greencheck & \blackcross & \greencheck\\
    Colmap-Free\citep{fu2024colmapfree3dgs}  & \blackcross & \greencheck & \blackcross & \greencheck\\
    \midrule
    \textbf{Ours} & \greencheck & \greencheck & \greencheck & \greencheck\\
    \bottomrule[1.5pt] 
    \end{tabular}}
    \caption{Summary of capabilities addressed in the previous models. In our work, the abbreviation for image deblurring models and view synthesis models are ``id'' and  ``vs'', respectively.}
    \label{tab:summary}
\end{table}

Recent research has aimed to reduce reliance on SfM techniques due to inaccurate and sparse Gaussian point initialization~\citep{foroutan2024evaluatingsfm, fu2024colmapfree3dgs}.
These approaches integrate pose estimation within the view synthesis frameworks, addressing joint photorealistic scene reconstruction and camera pose estimation. However, current methods are limited to specific scenarios, such as linear, slow-moving camera trajectories around a single object~\citep{xiang20233d, watson2022novel, lu20243d}, or address scenes with complex camera trajectories, yet typically rely on moderate to high-resolution inputs~\citep{muller2024multidiff,li2024spacetime, duan20244d}.
\Cref{tab:summary} summarizes the capabilities of existing methods.
Most focus on implicit neural field representations rather than direct camera pose optimization, presenting challenges for substantial camera movements.
Despite advancements, prolonged training times and diminished performance remain challenges in the joint novel view synthesis and camera pose estimation for multi-object blurry environments ~\citep{bian2023nope,badnerf}.

With this motivation, we aim to enhance the robustness and versatility of novel view synthesis in dynamically complex real-world scenarios while handling a variety of blur effects. To this end, we propose DBMovi-GS, a novel dynamic view synthesis method for blurry monocular videos.
DBMovi-GS is designed to handle three common types of blur encountered in real-world environments: camera motion blur, object motion blur, and defocus blur.
% Our method conducts a comprehensive analysis of Gaussian initialization to extract informative, dense 3D Gaussians from sparse points, which are utilized to model object motion and estimate camera poses, to performs photorealistic scene reconstruction in dynamic blurry real-world scenarios.
We conduct an in-depth analysis of Gaussian initialization to derive dense 3D Gaussians from sparse points. These Gaussians facilitate the robust modeling of object motion and camera pose, enabling photorealistic scene reconstruction in dynamic real-world scenarios.
Experimental results demonstrate that our model achieves superior performance in novel view synthesis and deblurring, effectively adapting to real-world scenes. Additionally, our work integrates seamlessly with rasterization techniques, enabling real-time rendering while preserving high-fidelity view synthesis.

\section{Related Work} \label{sec:background}
\subsection{Dynamic View Synthesis}
% The problem of novel view synthesis is strongly tied to multiview 3D reconstruction as it requires transporting pixels across views through the geometry of scenes.
% Dynamic View Synthesis aims to generate novel scenes with moving objects across viewpoints and timesteps.
Dynamic View Synthesis focuses on generating novel views of dynamic scenes, capturing both object motion and viewpoint changes across time.
While multi-view~\citep{bansal20204d, li2022neural}, stereo~\citep{attal2020matryodshka}, and sparse-view~\citep{irshad2023neo360} cameras have made progress, monocular videos remain challenging due to single-viewpoint ambiguities.
Earlier works utilize geometric priors, such as depth~\citep{du2021neural, park2024point} and optical flow~\citep{li2021neural, zhou2024dynpoint}, or
% conditions on position and view direction over time~\citep{li2022neural}, which requires a strong scene prior. 
% Several concurrent works employing
deformable radiance fields and regularization~\citep{luthra2024deblur, deblurnerf} to enhance reconstruction but often remain constrained to specific scenarios, limiting generalizability~\citep{li2024spacetime}.
% Another line of work achieves high-level view synthesis but struggles with blurry videos caused by object motion~\citep{you2024decoupling}.
Inspired by 3DGS, recent works achieve improved rendering quality~\citep{li2024spacetime, duan20244d} or represent dynamic scenes with optimized Gaussians~\citep{luiten2023dynamic, yang2023real}.
Another line of work focuses on camera pose estimation~\citep{fu2024colmapfree3dgs} or image deblurring~\citep{chen2024deblurgs, lee2024deblurring} separately, limiting their ability to achieve view synthesis with motion-related blur effects.
%  pixel-wise or
Building on prior work, our approach models both object and camera motion to achieve realistic view synthesis in challenging blurry environments.% However, most of the works assume dynamic scenes with a set of simplified 3D Gaussians and synthesis new views with a focus on reconstructing the entire scene or objects under simple frames.

\subsection{Deblurring 3D Representations}
Image deblurring has become a critical task, especially with the growing demand for high-quality visuals in real-world applications.
% variational Bayesian frameworks~\citep{huo2023blind, levin2011understanding, zhao2022deep} or
Recent methods leverage NeRF~\citep{mildenhall2021nerf,deblurnerf,bui2023dyblurf} and Gaussian splatting~\citep{chen2024deblurgs,lee2024deblurring} due to their effectiveness in scene reconstruction. 
NeRF-based approaches incorporate blur kernels with physical priors or model camera trajectories~\citep{badnerf} and voxel-based camera motions~\citep{lee2023exblurf}, showing efficiency in highly blurred images, but often rely on pretrained models~\citep{dpnerf} or lack 3D scene consistency~\citep{deblurnerf}.
% Recent advancements include camera motion trajectory estimation and voxel-based methods for complex camera motions~\citep{lee2023exblurf}.
Gaussian splatting-based approaches address blurry 3D scenes and probabilistic point rendering through camera motion trajectory modeling~\citep{lee2024deblurring,chen2024deblurgs,fu2024colmapfree3dgs}.
% \citet{lee2024deblurring} excels in addressing pixel-wise defocus or camera motion blur, but fails in modeling object motion-related blur. Another line of research tackles camera motion blur by interpolating camera poses over the shutter time~\citep{chen2024deblurgs,fu2024colmapfree3dgs}, which is effective in static scenes but less so for achieving fine detail of scene deblurring.
While prior works excel in addressing defocus blur~\citep{lee2024deblurring} or camera motion blur~\citep{chen2024deblurgs,fu2024colmapfree3dgs}, they often struggle with object motion blur and fine detail preservation. 
Our model effectively handles various blur types and improves the accuracy of object and pose estimation while preserving the fine scene details.

% Existing methods can be generally classified into two main categories. One formulates the deblurring problem as an optimization problem, where the latent sharp images and the blur kernel are jointly optimized using gradient descent.
% Another phase of the task is an end-to-end learning problem using deep convolution neural network techniques

% Deblur-NeRF \citep{deblurnerf} presents an end-to-end framework that jointly estimates pixel-wise spatially varying blur kernels along with the latent sharp radiance field but relies heavily on training deep neural networks, lacking geometric and appearance consistency in 3D scene representation. 
% To address these limitations, DP-NeRF \citep{dpnerf} introduces a novel blurring kernel incorporating two physical priors derived from the processes of blur acquisition and ray casting.
% BAD-NeRF \citep{badnerf} overcome these limitations by introducing a blurring kernel and modeling camera motion blur through camera motion trajectory estimation, necessitating a camera pose estimation method.
% Further advancing this line of work, ExBluRF \citep{lee2023exblurf} builds on BAD-NeRF by handling more complex camera motion trajectories and using a voxel-based NeRF, achieving greater efficiency and superior results, especially in highly blurred images.
% \input{sec/2_preliminaries}
\begin{figure*}[ht]
    \centering
    \includegraphics[width=1\linewidth]{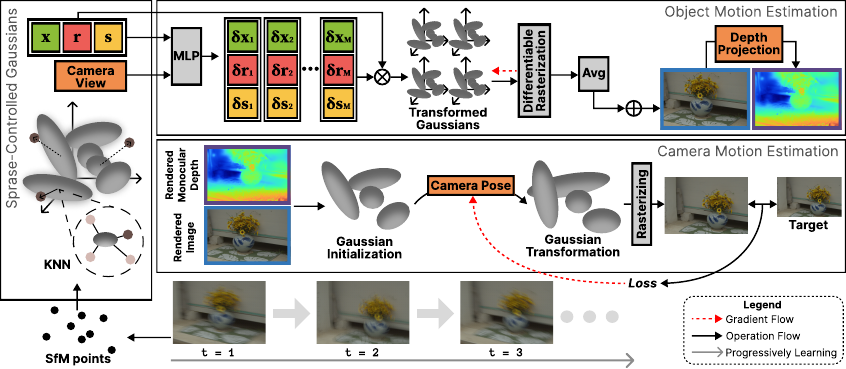}
    \caption{$\textbf{Overview}$.
    Given a sequence of unposed images and known camera intrinsics, our model recovers camera poses and synthesizes a sharp, photorealistic scene. We take a sequence of blurry images to learn sparse 3D Gaussian points, which are refined into dense Gaussian points via a K-Nearest Neighbor (KNN) search. These sparse-controlled 3D Gaussians are then used to estimate object motion and optimize camera poses via Gaussian transformations and color-consistent rendering. By minimizing photometric loss and positional loss during rendering, the model iteratively refines the 3D Gaussian representations as the camera moves.
    }
    \label{fig:architecture}
\end{figure*}
\section{Method} \label{sec:method}
DBMovi-GS aims to reconstruct dynamic scenes from blurry monocular videos.
We detail our approach for sparse-controlled Gaussian Splatting, followed by techniques for modeling object and camera motion.
Lastly, we describe how our model achieves dynamic novel view synthesis through a progressive learning framework.
An overview of our model is illustrated in~\cref{fig:architecture}.

\subsection{Sparse-Controlled Gaussian Extraction} \label{m:sparse-controlled}
Sparse-Controlled Gaussian extraction module involves augmenting the original \textit{sparse} 3D Gaussians to a set of \textit{dense} 3D Gaussians.
For each sparse point $\mathcal{P}_\text{sparse}$ computed from SfM, we employ KNN~\citep{knn} to identify the $K$ nearest control points and generate additional $\mathcal{N}_p$ Gaussian points, sampled randomly within the minimum and maximum value of the position of each existing point clouds. These boundaries determine the extent to which new dense Gaussians can be distributed around each original point.
% Unlike prior work~\citep{lee2024deblurring} that selectively prunes Gaussians near the far plane based on relative depth with varying threshold scalers, our method discards points solely based on distance thresholds.
We remove points exceeding a predefined nearest-neighbor distance threshold, $t_d$, effectively controlling point density and preventing excessive allocations or overestimations.
The newly generated points $\mathcal{P}_\text{KNN}$ are assigned color based on the nearest neighboring sparse 3D Gaussians.
The final set of \textit{dense} Gaussians is obtained by combining the generated point clouds with the original sparse points: $\mathcal{P}_\text{dense} = \mathcal{P}_\text{sparse} + \mathcal{P}_\text{KNN}$. Then, using the updated Gaussians, we perform two variety of motion estimations.
% $\mathcal{P}_\text{dense}$, is the union of $\mathcal{P}_\text{sparse}$ and $\mathcal{P}_\text{KNN}$.
% \begin{equation}
%     d(\mathbf{x}_i, \mathbf{x}_j) = \sqrt{\Sigma_{l=1}^{n} (x_{i,l} - x_{j,l})^2}
% \end{equation}
% where $x_{i,l}$ and $x_{j,l}$ are the $l$-th components of $\mathbf{x}_i$ and $\mathbf{x}_j$, respectively.

\subsection{Object Motion Estimation with Deblurring} \label{m:object-motion}
Object Motion blur arises when objects move through space at velocities exceeding the human eye to capture full visual fidelity.
% is a naturally occurring visual phenomenon that 
Accordingly, object motion estimation involves modeling actual physical movements to learn the transformations that modify its geometry.
% of the processed 3D Gaussians $\mathcal{P}_\text{dense}$

Similar to ~\citet{lee2024deblurring}, we produce $M$ additional sets of Gaussians by introducing small positional offsets to each dense Gaussian points $\mathcal{P}_\text{dense}$.
We then employ an MLP that take center position $\mu_j$, rotation $r_j$, scale $s_j$, and camera view $p$, along with its positional encoding $\gamma$, which outputs the transformed scaling factors of the $\textit{i}th$ predicted Gaussian points as $\{(\delta \mu_{ji}, \delta r_{ji}, \delta s_{ji})\}_{i=1}^M = \mathcal{F}_\theta (\gamma(\mu_j), r_j, s_j, \gamma(p))$.
% transformed position $\delta x_i$, rotation $\delta r_i$, and scaling $\delta s_i$
% that is shifted and rescaled for optimization stability. 
\iffalse
\begin{equation}
    \{(\delta \mu_j, \delta r_j, \delta s_j)\}_{i=1}^M = \mathcal{F}_\theta (\gamma(\mu_j), r_j, s_j, \gamma(p)).
    \label{eq:blur-transformation}
\end{equation}
\begin{equation}
    \{(\delta \mu_{ji}, \delta r_{ji}, \delta s_{ji})\}_{i=1}^M = \mathcal{F}_\theta (\gamma(\mu_j), r_j, s_j, \gamma(p)).
    \label{eq:blur-transformation}
\end{equation}
\fi
% where $\delta x_i$ is the \textit{i}th predicted position offset with scaled rotation $\delta r_i$ and scale factor $\delta s_i$.

The scaling factors are refined to obtain the transformed 3D Gaussians $G^*(\hat \mu_j,\hat r_j,\hat s_j)$ that model the blurriness caused by dynamic object motions as follows:
% , where $N_{G*}$ is the total number of dense 3D Gaussians generated.
\begin{subequations}
    \begin{align}
    \hat{\mu}_{ji} &= \mu_j + \lambda_p \delta \mu_{ji}, \\
    \hat{r}_{ji} &= r_j \cdot \rho_r \cdot \min(1.0, \lambda_q \delta r_{ji} + (1-\lambda_q)) \\
    \hat{s}_{ji} &= s_j \cdot \rho_s \cdot \min(1.0, \lambda_q \delta s_{ji} + (1-\lambda_q)),
    \end{align}
    \label{eq:gs-transformation}
\end{subequations}
where each set represents to 3D Gaussians from different viewpoints and is optimized during training to model blurriness caused by dynamic object movements.

The transformed Gaussians $G^*(\hat \mu_j, \hat r_j, \hat s_j)$ model motion-blurred scenes more flexibly as each Gaussian position, rotation, and scale factor are estimated.
% render them inadequate for effectively capturing detailed information from the scene.
From a practical perspective, defocus blur often occurs alongside motion blur due to the spatial characteristics of the environment. Therefore, the scaling factors are essential to reduce blurriness and improve motion estimation performance.
% , where information from a wide range of neighboring points is integrated.
% These parameters play a crucial role in reducing noise caused by object motion while simultaneously addressing defocus deblurring.

We render $M$ clean images from $M$ distinct input views of dense 3D Gaussians $G^*$, which are then averaged to generate object motion deblurred images $I$ using a differentiable rasterizer $\mathcal{R}$.
% \begin{equation}
%     I = \frac{1}{M}\sum_{i=1}^M I^i_\text{pred}.
%     \label{eq:image-pixel}
% \end{equation}
% \\As $G(\mu_j, r_j, s_j)$ learns the latent clean image,
During inference, no additional scaling factors are used for image rendering, thus, $\mathcal{F}_\theta$ is not activated, enabling real-time rendering of sharp images.
% where $I_{pred}$ is a clean image generted by the predicted scaling factors, aka the deltas.
% In addition to the rasterized clean image,
In the subsequent step, our method leverages the depth information derived from the rendered scene to estimate the camera motion. The transformed attributes are utilized to compute the depth of the rendered 3D Gaussian scene as:
\begin{equation}
    \hat D = \sum_{i \in \mathcal{N}} d_i \alpha_i T_i \quad \text{with} \quad T_i = \prod^{i-1}_{j=1}(1-\alpha_j),
    \label{eq:depth-pixel}
\end{equation}
where $d_i$ is the depth of the mean of the Gaussian point, which is the $\textit{z}$-coordinate of the position projected into the image view space as $d_i = (r_i p_i + T_i)_\textit{z}$.
% scaling factor 설명 추가할지말지

% $\textit{Remark:}$ 
% In this context, the parameters $\rho_s$ and, to a lesser extent, $\rho_r$, are employed to mitigate defocus blur.
% while concurrently tracking object motion.

% \begin{equation}
%     G_t^* = \text{argmin}_{c_t,r_t,s_t,a_t} \mathcal{L}_\text{image}(\mathbf{I}_b, \mathbf{I}_t)
% \end{equation}
% where $\mathcal{R}$ is the 3DGS render process.

\subsection{Camera Motion Estimation} \label{m:camera-motion}
We initialize 3DGS using the estimated rendered view and its monocular depth.
Following \citet{fu2024colmapfree3dgs}, the mean $\mu$ of 3D Gaussians is projected onto the 2D image plane as $\mu_\text{2D} = {\mathrm{K}(\mathrm{W}\mu)} / {(\mathrm{W}\mu)_z}$, 
where $\mathrm{K}$ represents the intrinsic projection matrix and $\mathrm{W}$ the viewing transformation matrix.
Consequently, estimating the camera pose is equivalent to determining the transformation of the 3D rigid Gaussians.

The pretrained Gaussians $G_t^*$ are transformed by a learnable $SE(3)$ transformation into the subsequent frame $G^*_{t+1}$, as $G^*_{t+1}=\mathbf{T}_t \odot G_t^*$, with an assumption that the camera moves in an arbitrary but smooth trajectory.
Mathematically, the optimization can be formulated as:
\begin{equation}
    \Theta^*, \Pi^* = \text{argmin}_{\Theta,\Pi} \mathcal{L}_\text{image}(\mathcal{R}(G^*_{t+1}), I_{t+1}),
    % \mathbf{T} = \text{argmin}_{\mathbf{T}_t} \mathcal{L}_\text{image}(\mathcal{R}(G^*_{t+1}), I_{t+1}).
\end{equation}
which is trained to minimize the photometric loss between the rendered image $G^*_{t+1}$ and the next frame $I_{t+1}$ with the camera parameters $\Pi$ and 3D Gaussian parameters $\Theta$ that are updated during optimization.
During optimization, we freeze all attributes of the pretrained 3D Gaussians.
% During optimization, we freeze all attributes of the pretrained 3D Gaussians $G_t^*$, effectively decoupling the camera pose from any deformation, densification, pruning, or self-rotation of the 3D Gaussian points.

% The pretrained Gaussians $G_t^*$ are transformed by a learnable $SE(3)$ transformation into the subsequent frame, denoted as $G^\ast_{t+1}=\mathbf{T}_t \odot G^\ast_t$ with an assumption that the camera moves in an arbitrary but smooth trajectory.
% The training objective is to minimize the photometric loss between the rendered image $G^*_{t+1}$ and the next frame $I_{t+1}$ by optimizing the transformation $\mathbf{T}_t$ for each time step.
% It is noteworthy that we freeze all attributes of the pretrained 3D Gaussians $G_t^*$, effectively decoupling the camera pose from any deformation, densification, pruning, or self-rotation of the 3D Gaussian points:
% \begin{equation}
%     \mathbf{T}_t^\ast = \text{argmin}_{\mathbf{T}_t} \mathcal{L}_\text{images}\big(\mathcal{R}(\mathbf{T}_t \odot G_t^\ast), I_{t+1})\big).
% \end{equation}
% The transformation $\mathbf{T}_i = [q_i | t_i]$ comprises separable parameters, a quaternion rotation $\mathbf{q}_i \in \text{SO}(3)$ and a translation $\mathbf{t}_i \in \mathbb{R}^3$.

\begin{table*}[ht]
    \centering
    \setlength{\tabcolsep}{0.5mm}
    \resizebox{1\textwidth}{!}{%
    \begin{tabular}{ccccccccccccccccccccccc}
    \toprule[1.5pt]
    \multicolumn{1}{c}{}&
    \multicolumn{1}{c}{}&
    \multicolumn{3}{c}{Camellia} &
    \multicolumn{3}{c}{Bench} &
    \multicolumn{3}{c}{Dragon} &
    \multicolumn{3}{c}{Sunflowers} &
    \multicolumn{3}{c}{Jars1} &
    \multicolumn{3}{c}{Jars2} \\
    \cmidrule(rl){3-5} \cmidrule(rl){6-8} \cmidrule(rl){9-11} \cmidrule(rl){12-14} \cmidrule(rl){15-17} \cmidrule(rl){18-20}
    \textbf{Type}&\textbf{Model} &
    {PSNR↑}&{SSIM↑}&{LPIPS↓}&{PSNR↑}&{SSIM↑}&{LPIPS↓}&{PSNR↑}&{SSIM↑}&{LPIPS↓}&{PSNR↑}&{SSIM↑}&{LPIPS↓} &
    {PSNR↑}&{SSIM↑}&{LPIPS↓}&{PSNR↑}&{SSIM↑}&{LPIPS↓} \\
    \midrule[1pt]
    id&Deblur-GS\citep{chen2024deblurgs} &\textbf{26.67}&\textbf{0.72}&\textbf{0.31}&\underline{26.89}&\underline{0.74}&0.29&27.40&0.57&\underline{0.54}&26.87&\underline{0.77}&0.41&24.49&0.63&\textbf{0.34}&\underline{23.46}&0.69&\textbf{0.34}\\
    id&Deblurring 3DGS\citep{lee2024deblurring} &24.92&0.56&0.53&25.15&0.57&0.59&\underline{29.05}&0.56&0.65&\underline{27.45}&0.73&\underline{0.39}&\textbf{25.59}&0.60&\underline{0.44}&22.72&0.57&0.52 \\
    vs&NoPe-NeRF\citep{bian2023nope} &23.45&0.57&0.66&23.05&0.64&0.66&28.68&0.61&0.69&26.05&0.73&0.59&24.14&0.63&0.64&22.25&0.67&0.63 \\
    vs&GS on the Move\citep{gsonthemove} &23.50&0.65&0.58&24.46&0.73&0.58&29.29&0.71&0.66&25.67&\underline{0.77}&0.50&24.46&\underline{0.69}&0.59&22.46&0.72&0.58\\
    vs&COLMAP-Free\citep{fu2024colmapfree3dgs} &19.60&0.69&0.63&21.72&4.46&\textbf{0.17}&25.79&\underline{0.83}&0.56&22.90&0.73&0.51&24.86&\textbf{0.78}&0.51&22.37&\underline{0.71}&0.53\\
    \midrule
    id+vs&\textbf{Ours} &\underline{25.98}&\underline{0.70}&\underline{0.39}&\textbf{26.99}&\textbf{0.75}&\underline{0.24}&\textbf{29.31}&\textbf{0.84}&\textbf{0.52}&\textbf{28.51}&\textbf{0.82}&\textbf{0.33}&\underline{24.95}&\textbf{0.78}&0.50&\textbf{24.11}&\textbf{0.74}&\underline{0.48}\\
    \bottomrule[1.5pt]
    \end{tabular}}
    \caption{Quantitative comparison of dynamic view synthesis on ExBluRF's real dataset.}
    \label{tab:result-exblurf}
% \end{table*}
% \begin{table*}[ht]
    \vspace{3mm}
    \centering
    \setlength{\tabcolsep}{0.5mm}
    \resizebox{1\textwidth}{!}{%
    \begin{tabular}{ccccccccccccccccccccccc}
    \toprule[1.5pt]
    \multicolumn{1}{c}{}&
    \multicolumn{1}{c}{}&
    \multicolumn{3}{c}{Church} &
    \multicolumn{3}{c}{Barn} &
    \multicolumn{3}{c}{Museum} &
    \multicolumn{3}{c}{Horse} &
    \multicolumn{3}{c}{Ballroom} &
    \multicolumn{3}{c}{Francis} \\
    \cmidrule(rl){3-5} \cmidrule(rl){6-8} \cmidrule(rl){9-11} \cmidrule(rl){12-14} \cmidrule(rl){15-17} \cmidrule(rl){18-20}
    \textbf{Type}&\textbf{Model} &
    {PSNR↑}&{SSIM↑}&{LPIPS↓}&{PSNR↑}&{SSIM↑}&{LPIPS↓}&{PSNR↑}&{SSIM↑}&{LPIPS↓}&{PSNR↑}&{SSIM↑}&{LPIPS↓} &
    {PSNR↑}&{SSIM↑}&{LPIPS↓}&{PSNR↑}&{SSIM↑}&{LPIPS↓} \\
    \midrule[1pt]
    id&Deblur-GS\citep{chen2024deblurgs} &25.19&0.82&0.19&24.29&0.69&0.34&22.80&0.67&0.29&20.43&0.65&0.39&26.12&0.85&0.13&28.06&0.82&0.27 \\
    id&Deblurring 3DGS\citep{lee2024deblurring} &22.71&0.75&0.19&28.01&\textbf{0.91}&\textbf{0.09}&33.84&0.95&\textbf{0.02}&27.36&0.88&\underline{0.07}&31.97&0.95&\textbf{0.02}&\textbf{33.76}&\underline{0.92}&\underline{0.08} \\
    vs&$\text{NoPe-NeRF}^\dagger$\citep{bian2023nope} &25.17&0.73&0.39&26.35&0.69&0.44&26.77&0.76&0.35&27.64&0.84&0.26&25.33&0.72&0.38&23.96&0.61&0.47 \\
    vs&GS on the Move\citep{gsonthemove} &30.62&0.92&\underline{0.07}&29.27&0.97&0.06&35.71&0.97&\textbf{0.02}&24.48&0.94&0.10&\textbf{35.53}&\textbf{0.97}&\textbf{0.02}&32.41&0.94&0.10 \\
    vs&COLMAP-Free\citep{fu2024colmapfree3dgs} &30.38&\underline{0.93}& \underline{0.09}&\textbf{30.43}&\underline{0.89}&\underline{0.11}&\underline{29.50}&0.90&0.10&\underline{34.10}&\underline{0.96}&\textbf{0.05}&32.64&\underline{0.96}&\underline{0.05}&32.81&\underline{0.92}&0.14  \\
    \midrule
    id+vs&\textbf{Ours} &\textbf{34.04}&\textbf{0.98}& \textbf{0.02}&\textbf{30.43}&\underline{0.89}&\textbf{0.09}&\textbf{29.80}&\textbf{0.98}&\underline{0.08}&
    \textbf{36.52}&\textbf{0.98}&\textbf{0.014}&\underline{35.01}&\textbf{0.97}&\textbf{0.02}&\underline{33.20}&\textbf{0.93}&\textbf{0.06}\\
    \bottomrule[1.5pt]
    \end{tabular}}
    \caption{Quantitative comparison of dynamic view synthesis on Tanks and Temples' real dataset. $\dagger$: Results from paper.}
    \label{tab:result-tank}
% \end{table*}
% \begin{table*}[h]
    \vspace{3mm}
    \centering
    \setlength{\tabcolsep}{1mm}
    \resizebox{1\textwidth}{!}{%
    \begin{tabular}{ccccccccccccccccccccccc}
    \toprule[1.5pt]
    \multicolumn{1}{c}{} & 
    \multicolumn{1}{c}{} & 
    \multicolumn{3}{c}{Church} &
    \multicolumn{3}{c}{Barn} &
    \multicolumn{3}{c}{Museum} &
    \multicolumn{3}{c}{Horse} &
    \multicolumn{3}{c}{Ballroom} &
    \multicolumn{3}{c}{Francis} \\
    \cmidrule(rl){3-5} \cmidrule(rl){6-8} \cmidrule(rl){9-11} \cmidrule(rl){12-14} \cmidrule(rl){15-17} \cmidrule(rl){18-20}
    \textbf{Type} & \textbf{Model} &
    {$\text{RPE}_t$↓} & {$\text{RPE}_r$↓} & {$\text{ATE}$↓} & {$\text{RPE}_t$↓} & {$\text{RPE}_r$↓} & {$\text{ATE}$↓} & {$\text{RPE}_t$↓} & {$\text{RPE}_r$↓} & {$\text{ATE}$↓} & {$\text{RPE}_t$↓} & {$\text{RPE}_r$↓} & {$\text{ATE}$↓} & {$\text{RPE}_t$↓} & {$\text{RPE}_r$↓} & {$\text{ATE}$↓} & {$\text{RPE}_t$↓} & {$\text{RPE}_r$↓} & {$\text{ATE}$↓} \\
    \midrule[1pt]
    vs&$\text{NoPe-NeRF}^\dagger$\citep{bian2023nope}&0.034&0.008&0.008 & 0.0046 & 0.032& 0.004 & 0.207 & 0.202 & 0.020 & 0.179 & 0.017 & 0.003 & 0.041 & 0.018 & 0.002 & 0.057 & 0.009 & 0.005 \\
    vs&COLMAP-Free\citep{fu2024colmapfree3dgs} &0.012&0.018&0.002&0.040&0.034&0.004&0.052&0.215&0.004&0.112&0.057&0.003&0.037&0.017&0.003&0.029&0.154&0.006\\
    \midrule
    id+vs&\textbf{Ours} &0.011&0.018&0.002&0.039&0.036&0.004&0.051&0.202&0.004&0.101&0.046&0.001&0.030&0.009&0.002&0.025&0.102&0.006\\
    \bottomrule[1.5pt]
    \end{tabular}}
    \caption{Quantitative results of camera pose estimation on Tanks and Temples. We exclude results from GS on the Move, as this model primarily targets smartphone data or basic synthetic datasets and does not record external pose-related sources. $\dagger$: Results from paper.}
    \label{tab:result-camera-pose-2}
\end{table*}

\subsection{Model Training} \label{m:optimization}
We optimize our approach with the following objective:
\begin{subequations}
    \begin{align}
    \mathcal{L} &= \mathcal{L}_\text{image} + \lambda_\text{depth}\mathcal{L}_\text{depth} +
    \lambda_\text{pose}\mathcal{L}_\text{pose}, \\
    \mathcal{L}_\text{image} &= \lambda \mathcal{L}_1 + (1-\lambda)\mathcal{L}_\text{D-SSIM}, \\
    \mathcal{L}_\text{pose} &= \| \text{max}( s_j,\epsilon_\text{pose}) \|_2
    \end{align}
\end{subequations}
where $\lambda_\text{depth} = 0.01$ and $\lambda_\text{pose} = 1$.
% $\lambda_\text{depth} = 0.01$
% Depth loss is formulated as the L1 distance, guiding the rendered depth toward the estimated depth.
The image loss combines L1 and D-SSIM terms, weighted by $\lambda = 0.2$.
The pose loss regularizes the position scale of each Gaussian using its scaling factor, with $\epsilon_\text{pose}$ controlling tolerance to minor localization errors in the original Gaussian points.
Please refer to the supplementary materials for full details.

% In order to derive each loss term,
% \begin{equation}
%     \frac{\delta \mathcal{L}}{\delta \mathbf{T}} = \frac{\delta \mathcal{L}}{\delta \mathbf{\hat m}} \frac{\delta \mathbf{\hat m}}{\delta \mathbf{T}}, \quad 
%     \text{where } \frac{\delta \mathcal{L}}{\delta \mathbf{T}} =\frac{\delta \mathcal{L}}{\delta \mathbf{T}} \frac{\delta \mathcal{L}}{\delta \mathbf{T}} 
% \end{equation}
% Following the principle of being as rigid as possible \citep{sorkine2007arap}, which enables the deformation constrained to be as rigid as possible at a local level, effectively reducing artifacts and maintaining the geometric integrity of the shape while allowing for necessary transformations.

\section{Experiment} \label{sec:experiment}
% \subsection{Dataset and Baseline}
In this section, we evaluate our experiments using the ExBluRF~\citep{lee2023exblurf} and Tanks and Temples~\citep{tanks} real camera motion datasets, comparing it against five models listed in ~\cref{tab:summary}.
% \textbf{DeblurNeRF}~\citep{deblurnerf} real camera motion blur dataset consists of real-life scenes where multi-view cameras are manually positioned to replicate real data-capture scenarios, assuming linear camera motion in constant velocity.
% Each scene in \textbf{ExBluRF}~\citep{lee2023exblurf} real dataset comprises 29 blurry training images and 5 sharp test images, incorporating random 6-DOF camera motion trajectories.
% \textbf{Tanks and Temples}~\citep{tanks} real camera motion dataset assesses the quality of novel view synthesis and pose estimation across 6 scenes varying from indoor to outdoor environments.
% Details of each dataset are provided in the supplementary materials.
% We evaluate novel view synthesis and camera pose estimation using the following evaluation metrics.
For novel view synthesis evaluation, we report the standard metrics: PSNR~\citep{psnr}, SSIM~\citep{ssim}, and LPIPS~\citep{lpips}.
For camera pose estimation, we measure Absolute Trajectory Error (ATE) and Relative Pose Error (RPE), in terms of their rotation (r) and translation (t).

\subsection{Results on Dynamic View Synthesis}
% \paragrapht{Deblur-NeRF.}
% Our method achieves comparable performance as shown in~\cref{tab:result-realcam}.
% % Since the dataset is specifically designed for image deblurring tasks, ``id'' models naturally achieve the best performance, particularly as it targets significant blur effects. ``vs'' models, such as GS on the Move and COLMAP-Free, exhibit lower performance, likely due to their limited ability to effectively handle deblurring.
% As the Deblur-NeRF dataset is tailored for deblurring tasks, the ``id'' models excel at addressing the blur effects, whereas the ``vs'' models exhibit diminished performance, likely due to their limited ability to handle scene blurriness. In contrast, our model achieves comparable results for dynamic view synthesis by effectively balancing image deblurring and novel view synthesis, all while maintaining real-time rendering capabilities.
% % Importantly, our approach preserves real-time rendering capabilities, achieving a high FPS, in contrast to ``id'' models that face limitations in this regard.
% % \cref{fig:dvs-exblurf} presents the qualitative results on the real dataset, showcasing our ability to produce sharp, fine details that baselines fail to reconstruct.
As illustrated in~\cref{tab:result-exblurf}, our method achieves accurate view synthesis from blurry monocular video on the ExBluRF dataset. In contrast, baselines struggle to handle scene blurriness, often producing inaccurate sharp reconstructions, particularly in densely clustered regions where occluded or ambiguous objects overlap.
Furthermore, our method outperforms existing models in the Tanks and Temples dataset, as demonstrated in~\cref{tab:result-tank}. The dataset's diversity of indoor and outdoor environments highlights the robustness of our approach in synthesizing complex scenes.
% without constraints.
% Deblur-GS is the current state-of-the-art and our model achieves comparable performance.
% Our method achieves performance superior to previous ``vs'' models as demonstrated in~\cref{tab:result-tank}. The dataset diversity in indoor and outdoor environments highlights our model's robustness in handling complex scenes without constraints.
% The rendering results, along with its corresponding 3D point cloud, are presented in part of~\cref{fig:cam-pose}.
% We omit 3DGS+COLMAP, as~\citet{fu2024colmapfree3dgs} provides an in-depth ablation of COLMAP-Free and 3DGS+COLMAP, demonstrating superior performance over original 3DGS.
% Qualitative results are available in the supplementary materials.
% As expected, ``id'' models demonstrate suboptimal performance due to their limited capacity to address complex camera movements.
% making it less dependent on deblurring effects.

\subsection{Results on Camera Motion Estimation}
Following ~\citet{fu2024colmapfree3dgs}, we refine the estimated camera poses using Procrustes analysis and compared them with the ground-truth poses from the training views. Experimental results demonstrate that jointly handling image deblurring and pose optimization maintains performance and, in fact, surpasses baselines on the Tank and Temples in~\cref{tab:result-camera-pose-2}. 
% Qualitative results, including rendered pose trajectories, are provided in the supplementary material.
% We posit that the observed discrepancies in terms of RPEr and RPEt likely arise from the reliance on photometric loss for relative pose estimation in local regions.
% The exclusion of point cloud loss results in a notable decline in pose estimation accuracy~\citep{bian2023nope}.
 % integrates additional constraints on relative poses, such as the chamfer distance between two point clouds, in which

Due to space constraints, qualitative results for dynamic view synthesis and camera motion estimation, including pose trajectory, are provided in the supplementary material.

%%%%%%%%%%%%%%%%%%%%%%%%%%%%%%%%%%%%%%%%%%%%%%%%%%%%%%%%%%%%%%%%%%%%%%%%%%%%%%%%%%%%%%%%%%%%%%%%%%%%%%%%%%%%%%%%%%%%%%%%
% \begin{table}[t]
% \centering
%     \setlength{\tabcolsep}{1.2mm}
%     \resizebox{1\columnwidth}{!}{%
%     \begin{tabular}{@{}lccc|c@{}}
%     \toprule[1.5pt]
%     & \multicolumn{1}{c}{dataset 1} & \multicolumn{1}{c}{dataset 2} & \multicolumn{1}{c}{dataset 3} & \multicolumn{1}{c}{\begin{tabular}[c]{@{}c@{}}Training\\ Time \end{tabular}} \\
%     \midrule[1pt]
%     Deblur-GS\citep{chen2024deblurgs} & 10000 & 10000 & 10000 & $\times$ 0.80 \\
%     COLMAP-Free\citep{fu2024colmapfree3dgs} & 12000 & 12000 & 12000 & $\times$ 0.95 \\
%     \midrule
%     \textbf{Ours} & 15000 & 15000 & 15000 & - \\
%     \bottomrule[1.5pt]
%     \end{tabular}}
%     \caption{Evaluation of point addition on Tank's real dataset. Time difference indicates the difference in training time compared to Ours.}
%     \label{tab:point-addition}
% \end{table}
\section{Conclusion} \label{sec:conclusion}
In this paper, we present DBMovi-GS, a method for dynamic view synthesis from blurry monocular video, addressing the limitations of prior approaches in handling dynamic objects.
By densifying the initial point clouds and employing Gaussian representations to jointly model object and camera motion, our method achieves highly detailed, consistent, and realistic reconstructions of dynamic scenes.

\newpage
{
    \small
    \bibliographystyle{ieeenat_fullname}
    \bibliography{main}
}

\clearpage
\setcounter{page}{1}
\maketitlesupplementary

% \subsection{Hyperparameter}
% \begin{table}[ht]
%     \centering
%     \begin{tabular}{ccccc}
%     \toprule[1.5pt]
%        a  & a & ExBlurf & Real Cam & Tank \\
%     \midrule[1pt]
%        batch size  &  &  &  & \\
%        learning rate  &  &  &  & \\
%        learning rate  &  &  &  & \\
%        \bottomrule[1.5pt]
%     \end{tabular}
%     \caption{Caption}
%     \label{tab:my_label}
% \end{table}

Within the supplementary material, we provide:
\begin{itemize}
    \item Pseudocode of our proposed Sparse-Controlled Gaussian Extraction framework in~\cref{supp:pseudocode}.
    \item Implementation details, including hyperparameters and training details in~\cref{supp:model_details}.
    \item Further description of the datasets and baselines, including training specifics, in ~\cref{supp:dataset_details,supp:baseline_details}, respectively.
    % \item Baseline details and preprocessing requirements for evaluation, , in~\cref{supp:baseline_details}.
    \item Detailed procedure of our proposed method in ~\cref{supp:method}
    \item Qualitative results for dynamic view synthesis and camera pose estimation in ~\cref{supp:results}.
    \item Ablation studies on the loss function and the Gaussian splats in ~\cref{supp:ablation}
\end{itemize}

\section{Sparse-Controlled Gaussian Pseudocode} \label{supp:pseudocode}
\begin{algorithm}
\caption{Sparse-Controlled Gaussian Initialization}
\label{alg:algorithm}

\begin{algorithmic}[1]
% $G_\text{sparse} = \{G_i\}_{i=1}^{N}$
    \Require Sparse points $\mathcal{P}_\text{sparse}$
    \Require Number of new points $N_p$
    \Require KNN parameter $K$
    \Require Distance threshold $t_d$
    \Ensure Dense points $\mathcal{P}_\text{dense}$
    % \State Initialize $\mathcal{P}_\text{dense} \gets \mathcal{P}_\text{sparse}$
    \For{each point $p \in \mathcal{P}_\text{sparse}$}
        \State Uniformly sample $\mathcal{N}_p$ new points to form $\mathcal{P}_\text{new}$.
        % within bounds of $\mathcal{P}_\text{sparse}$ to form $\mathcal{P}_\text{new}$
        \For{each point $p_\text{new} \in \mathcal{P}_\text{new}$}
            \State Identify $K$ nearest neighbors $\mathcal{N}_K$ from $\mathcal{P}_\text{sparse}$
            \State Compute distances $d(p_\text{new}, \mathcal{N}_K)$
            \If{$\min d(p_\text{new}, \mathcal{N}_K) > t_d$}
                \State Discard $p_\text{new}$
            \Else
                \State Assign interpolated properties to $p_\text{new}$
                \State Add $p_\text{new}$ to $\mathcal{P}_\text{KNN}$
            \EndIf
        \EndFor
    \EndFor
    
    \State $\mathcal{P}_\text{dense} = \mathcal{P}_\text{sparse} + \mathcal{P}_\text{KNN}$ \\
    % \State $G_\text{dense} \gets G_\text{sparse} \cup G_\text{KNN}$ \Comment{Merge sparse and dense Gaussians}
    
    \Return $\mathcal{P}_\text{dense}$
\end{algorithmic}
\end{algorithm}

\section{Implementation Details} \label{supp:model_details}
\subsection{Motion Estimation}
The positional encoding utilized for the scaling factors during object motion estimation incorporates two key features: multi-frequency representation and dimensionality expansion.
We adopt a sinusoidal positional encoding following the approach in~\citet{lee2024deblurring}, employing frequencies of $2^k\pi$ and dimensionality of $2L$ for each encoding. This design enables the model to effectively capture coarse and fine-grained positional information, which is essential for accurately modeling object motion at varying scales and complexities.

Given the high proximity between consecutive frames, the associated transformations are relatively small, allowing for efficient optimization.
Additionally, the pose estimation step leverages this efficient encoding to achieve precise results within a runtime of less than a minute.
This streamlined process not only minimizes the computational overhead but also ensures scalabilities for processing large datasets with complex and dynamic motion patterns.
 % and the encoding is applied element-wise further increasing the output dimension.
% During the process of generating dense 3d Gaussian, we randomly sample additional Gaussians from a uniform distribution $U(a,b)$, where $a$ and $b$ are the boundaries of displacement for new samples surrounding each existing point cloud.

\section{Dataset Details} \label{supp:dataset_details}
\subsection{ExBluRF}
ExBluRF~\citep{lee2023exblurf} presents challenging motion-blurred images by incorporating random 6-DOF camera motion trajectories.
Each scene in this dataset comprises 29 blurry training images and 5 sharp test images.
The real datasets from ExBluRF were acquired using a dual-camera system; one camera captured a single blurry image with a long exposure time, while the other recorded a sequence of sharp images during the same exposure. A total of eight scenes were captured, with each scene containing between 20 to 40 multi-view blurry images along with their corresponding sequences of sharp images.

% \subsection{Deblur-NeRF}
% DeblurNeRF~\citep{deblurnerf} consists of diverse scenes and multi-view cameras are manually positioned to replicate real-world data-capture scenarios. Our work uses only the real camera motion blur. To render images with camera motion blur, the camera poses are randomly perturbed. Linear interpolation is then applied to the original and perturbed poses for each viewpoint. Images are rendered from these interpolated poses and subsequently blended in linear RGB space, resulting in the final set of blurry images.
\subsection{Tanks and Temples}
The Tanks and Temples~\citep{tanks} dataset assesses the quality of view synthesis and the accuracy of pose estimation, which include both indoor and outdoor environments. We use the real dataset version only and sample 7 images from every 8-frame clip for training per scene, reserving the remaining images to test the quality of view synthesis. Camera poses are estimated and evaluated using all training samples. \cref{tab:tank_scene} provides details about the video sequences for each scene in the Tanks and Temples benchmark.
\begin{table}[t]
    \centering
    \resizebox{1\columnwidth}{!}{%
    \begin{tabular}{cccc}
    \toprule[1.5pt]
       Scene  & Type & Frame rate (fps) & Max. rotation (deg) \\
    \midrule[1pt]
       Church  & indoor  & 30 & 37.3$^{\circ}$ \\
       Barn    & outdoor & 20 & 47.5$^{\circ}$ \\
       Museum  & indoor  & 20 & 76.2$^{\circ}$ \\
       Horse   & outdoor & 30 & 39.0$^{\circ}$ \\
       Ballroom& indoor  & 20 & 30.3$^{\circ}$ \\
       Francis & outdoor & 20 & 47.5$^{\circ}$ \\
       \bottomrule[1.5pt]
    \end{tabular}}
    \caption{Details of training sequences of Tanks and Temples dataset. Fps denotes frame per second per scene. Max. rotation denotes the maximum relative rotation angle, in degrees, between two frames in the selected sequence.}
    \label{tab:tank_scene}
\end{table}

% \begin{table}[ht]
%     \centering
%     \setlength{\tabcolsep}{1.2mm}
%     \resizebox{1\columnwidth}{!}{%
%     \begin{tabular}{ccccc}
%     \toprule[1.5pt]
%        Scenes  & Type & Seq. length & Frame rate & Max. rotation \\
%     \midrule[1pt]
%        Ball      & indoor  &  &  &  \\
%        Basket    & indoor &  &  &  \\
%        Buick     & indoor  &  &  &  \\
%        Coffee    & indoor &  &  &  \\
%        Decoration& indoor  &  &  &  \\
%        Girl      & indoor    &  &  &  \\
%        Heron     & indoor    &  &  &  \\
%        Puppet    & indoor    &  &  &  \\
%        Stair     & indoor    &  &  &  \\
%        \bottomrule[1.5pt]
%     \end{tabular}}
%     \caption{Details of selected sequences of Deblur-NeRF's dataset.}
%     \label{tab:deblur_scene}
% \end{table}

\section{Baseline Details} \label{supp:baseline_details}
We outline the preprocessing and setup procedures applied to the baseline models, including the additional adaptations required to align them with our experimental conditions.
Technical specifications are also detailed.

\subsection{Deblur-GS}
Deblur-GS~\citep{chen2024deblurgs} only supports $\texttt{SIMPLE\_PINHOLE}$ and does not support the $\texttt{SIMPLE\_RADIAL}$ model used in the Tanks dataset. Both refer to specific camera models used during 3D reconstruction, typically implemented in COLMAP or similar structure-from-motion tools.
When extracting camera poses from COLMAP, images are not transformed according to the strict distortion model; instead, the field of view (FOV) is utilized directly.
Following the configuration of NoPe-NeRF, we applied the same approach in Deblur-GS, calculating the field of view in the same manner as the specific camera model used during 3D reconstruction.

\subsection{Gaussian Splatting on the Move}
As outlined in the original paper~\citep{gsonthemove}, for the datasets lacking camera poses or 3D point cloud data (\ie, \text{transform.json} in the Nerfstudio format), we process raw benchmark input data using the provided Python script. Specifically, COLMAP was employed to reconstruct point clouds and generate camera poses from blurry image sequences (\ie., \text{sparse\_pc} and \text{transform.json}).
One explanation for the findings in the new study is that extracting feature points from blurry images often required multiple attempts to ensure successful reconstruction.
The training was conducted exclusively on blurry image samples, with sharp images reserved for evaluation to assess the prediction of sharp outputs.
Training for a single image sequence typically converged within 12000 epochs, with total training time generally under one hour per sequence, depending on the dataset.
At present, no code is provided for computing pose estimation errors, as the focus of the study is on smartphone data, with no intention to analyze the intricacies of pose estimation in this context.
The experimental setup was implemented on Ubuntu 20.04 using a Conda environment with Python 3.8.20, CUDA 12.1, and six RTX 3090 GPUs, each equipped with 24GB of VRAM.
% This dataset lacks an external position source as ground truth, making it impossible to calculate metrics such as Relative Pose Error (RPE) or Absolute Pose Error (APE) without millimeter-level ground truth.

\subsection{NoPe-NeRF}
For the NoPe-NeRF~\citep{bian2023nope} dataset, the experimental setup utilized Ubuntu 20.04, a Conda environment with Python 3.9.18, CUDA 12.1, and six RTX 3090 GPUs, each featuring 24GB of VRAM.
During preprocessing, depth maps were generated for each image within the provided image sequences.
The model included camera pose and point cloud reconstruction data, eliminating the need for external reconstruction methods.
Training times varied by dataset but generally required less than 1.5 days per scene.
The evaluation was conducted using sharp images to assess the predicted sharp outputs.
Early stopping was applied during training, with convergence typically achieved in fewer than 10000 epochs for each sequence.

\section{Method Details}\label{supp:method}
\subsection{Preliminaries}
\paragrapht{3D Gaussian Splatting.}
3DGS~\citep{gaussian-splatting} represents a volumetric scene as a set of 3D Gaussians.
A scene is parameterized as a set of Gaussian points with center position $\mu$, opacity $\sigma$, and sphere harmonics coefficient $sh$. 
Each 3D Gaussian is defined by a 3D covariance matrix and its center position $\mu$, which is referred to as the mean value of the Gaussian:
% $\mathcal{G} = \{ G_i:\mu_i, q_i, s_i, \sigma_i, sh_i | i=1,\dots,N\}$
\begin{equation}
\label{eq:gaussian-splatting}
    %G(x,r,s) = e^{-\frac{1}{2} x^T \sum^{-1}(r,s) x}.
    G(x,r,s) = \exp{\Big(-\frac{1}{2} x^T \Sigma^{-1}(r,s) x\Big)}.
\end{equation}

% Given camera pose $\mathbf{V}={R_t, t_t}$ and camera intrinsic matrix $\mathbf{U}$,
For differentiable optimization, the covariance matrix is decomposed into two learnable components as $\Sigma(r,s) = R S S^T R^T$, where $R$ is for rotation quaternion and $S$ is for scaling parameter.
These 3D Gaussians are projected into 2D image space to obtain a 2D covariance matrix in camera coordinates $\Sigma(r,s)'$ for rendering via viewing transformation matrix $W$ and Jacobian of the affine approximation of the projective transformation $J$:
\begin{equation}
    \Sigma(r,s)' = J W \Sigma(r,s) W^T J^T.
\end{equation}

After projection, each pixel is computed by blending the $N$ ordered points that overlap the 2D Gaussian pixel as:
% $\alpha = \Lambda e^{\frac{-1}{2}x^t\Sigma^{'-1}x}$. Then, the color of a pixel $u$ is rendered by performing $\alpha$-blending, where $c_i$ is the spherical function of its view direction:
\begin{equation}
    \hat C = \sum^{N}_{i=1} c_i \alpha_i T_i,
    \quad \text{where} \quad
    T_i = \prod^{i-1}_{j=1}(1-\alpha_j).
    \label{eq:color-pixel}
\end{equation}
where $c_i$ and $\alpha_i$ represent the color and density of each point computed by a 3D Gaussian with covariance matrix multiplied by per-point opacity and $\textit{sh}$ color coefficients.

\paragrapht{Image Deblurring.}  \label{m:blur}
Image blur is a common manifestation of camera shake or defocus, which occurs when the shutter speed is insufficient to maintain a stable camera pose or the light ray from a scene point forms a circle of confusion on the camera lens~\citep{machinevision}.
Image deblurring involves removing the blurring artifacts from images to restore the original content.
Among diverse applications of deep learning in deblurring, we address three types of blur frequently encountered in real-world scenarios~\citep{xiang2024application}.

\textit{Object motion blur} arises from the relative motion between an object and the camera system during exposure, which occurs when capturing fast-moving objects or with long exposure times. 
% typically 
\textit{Camera motion blur} is caused by the camera's motion during exposure, common in photography or low-light conditions, such as indoors or at night. This blur effect can be complex due to unpredictable hand movements, leading to in-plane or out-of-plane camera translations.
% and rotations.
\textit{Defocus blur} occurs when parts of a scene fall outside the camera's depth of field, the range where objects appear sharp.
Defocus blur is relatively easy to handle as it involves adjusting the geometry of each 3D Gaussian. In contrast, camera and object motion blur arise from natural or intentional physical movements.
% In our work, we do not explicitly train to reduce defocus blur as it is implicitly handled as part of object motion estimation, which adjusts scale and rotation factors that also contribute to image defocusness.

\subsection{Sparse Gaussian Extraction}
% To represent sparse 3D Gaussian points for learning motion dynamics under blurry inputs, 
% $\mathcal{P}_\text{add}={p_i \in \mathbb{R}^3}, i \in {1,2, \dots N_p}$
% which enhances the accuracy of motion dynamics modeling and defocus blur under blurry inputs.
% 3DGS~\citep{gaussian-splatting} construct a sparse set of 3D Gaussians with the corresponding calibrated camera poses (\textit{ref.}~\cref{m:gaussian}).
For blurry input images, 3DGS often produces large Gaussians or fails to capture distant points along the object edges, leading to a loss of essential details required for sharp image generation.
This ambiguity in Gaussian points,
% due to their reliance on high-quality point cloud initialization from COLMAP,
stemming from their dependence on high-quality point cloud initialization from COLMAP,
poses challenges in practical applications with large-scale, complex dynamic motions~\citep{foroutan2024evaluatingsfm}. 
Thus, it hinders a new approach for point initialization and extraction that captures useful information from blurry monocular images, thereby reducing reliance on COLMAP for photorealistic dynamic view synthesis.
% such as dynamic view synthesis from blurry monocular real-world videos.
% A common limitation in all models is that the aspect of object motion blur modeling is missing. in real-world environment, dynamic objects exists.

\subsection{Object Motion Estimation}
The object motion estimation is formulated as follows:
\begin{equation}
    I^i_\text{pred} = \mathcal{R}(\{G^*(\hat \mu_{ji}, \hat r_{ji}, \hat s_{ji})\}_{j \in \mathcal{P}_\text{dense}}),
\end{equation}
where the scaling factors are adjusted by parameter $\lambda$, clipped, and element-wise multiplied with the original Gaussian factors to obtain its transformed sets. Unlike previous methods that model object motion blur and defocus blur separately, our approach introduces an additional set of Gaussians with two learnable parameters, $\rho_r$ for rotation and $\rho_s$ for scale, enabling Gaussian covariance modeling based on scaling factors. These parameters play a pivotal role in simultaneously reducing noise from complex object motion and effectively addressing defocus deblurring.

\subsection{Camera Motion Estimation}
Camera motion blur, often resulting from camera shake, occurs when the shutter speed is insufficient to maintain a stable pose.
Recent approaches demonstrate that estimating camera poses and rendering images can be achieved simultaneously,
making pose estimation effectively equivalent to determining 3D rigid transformations ~\citep{bian2023nope,fu2024colmapfree3dgs}.
% thus, estimating camera poses is equivalent to determining the 3D rigid transformations. 
Leveraging this insight, our model explicitly approximates the camera motion trajectory and optimizes a global transformation for all Gaussians for each camera, without relying on SfM.

In general, the camera parameters include camera intrinsics, poses, and lens distortions.
In our work, we utilize camera poses only, such that camera pose for frame $I_i$ is a transformation $\mathbf{T}_i = [\mathbf{q}_i | \mathbf{t}_i]$ that comprises a rotation $\mathbf{q}_i \in \text{SO}(3)$ and a translation $\mathbf{t}_i \in \mathbb{R}^3$.

\subsection{Model Training}
\paragrapht{Progressively Learning.}
Applying the aforementioned technique to each pair of images determines the relative pose between the initial frame and any subsequent frame at timestep $t$. However, noise in the relative poses can hinder the optimization of the Gaussian models for the entire scene.
To address this issue, we follow~\citet{fu2024colmapfree3dgs} by estimating the relative pose between $I_\text{t+1}$ and $I_\text{t+2}$ when the next frame $I_\text{t+2}$ becomes available.
The global Gaussian points are updated over the whole iteration utilizing the estimated relative pose and the two observed frames.
The progressive learning approach continuously refines the scene by leveraging the sequential nature of video frames. The global points expand the 3D Gaussian representation, better capturing complex camera movements and enabling accurate pose estimation during novel view synthesis.

A fundamental hypothesis underlying our approach is that the scales should be consistent. While points are initialized at their centers and new points are added in proximity to the existing ones, these primitives may diverge from their parent points after the optimization.
% To mitigate this issue,
Although $\mathcal{L}_\text{image}$ alone yields comparable rendering results without the need for additional supervision, we leverage depth loss $\mathcal{L}_\text{depth}$ to minimize the prediction error associated with the depth estimation from the depth projector (\textit{ref.}~\cref{eq:depth-pixel}).
% from ~\cref{eq:depth-pixel}.

We employ the Adam optimizer with a learning rate of $1 \times 10^3$ for the MLP and $1.6 \times 10^4$ for 3D Gaussians.
For the real camera motion blur dataset, the pruning and densification thresholds for 3D Gaussians are set to $1 \times 10^{-2}$ and $5 \times 10^{-4}$, respectively.
The MLP consists of 4 layers, with the first three layers shared across tasks.
These layers extract features that are subsequently fed into three separate output heads, which predict $\delta \mu$, $\delta r$, and $\delta s$ scaling factors.
Each layer comprises 64 hidden units, utilizing ReLU activations, and is initialized via Xavier initialization \citep{xavier}.
 The regularization parameters, $\lambda_p$ and $\lambda_s$, are both set to $1 \times 10^{-2}$.
To address point cloud sparsity, additional points are introduced starting from the 2000$th$ iteration.
The number of nearest neighbors is set to 4, with a minimum distance threshold set to 2.
Depth-based pruning is performed using a multiplier of 3.
For camera motion deblurring, the hyperparameter $M$ is set to 4.
Overall, the model is trained for an average of 20000 iterations.
% , with the number of added points capped at 200000 based on point size.

\begin{figure}[ht]
    \centering
    \includegraphics[width=1\linewidth]{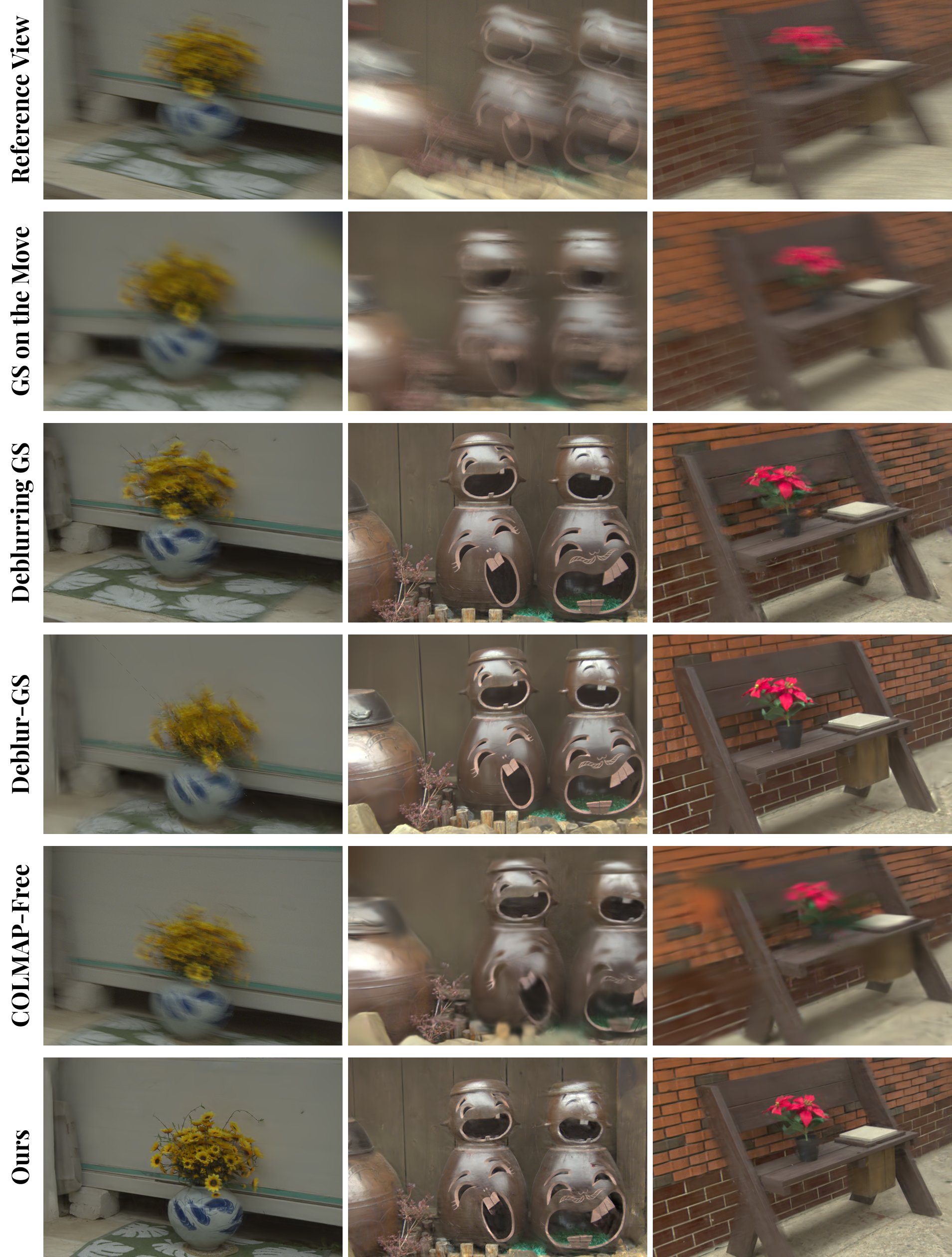}
    \caption{Qualitative comparison of dynamic view synthesis on ExBluRF dataset. We compare our approach against Gaussian-based methods, including GS on the Move\citep{gsonthemove}, Deblurring 3DGS\citep{lee2024deblurring}, Deblur-GS\citep{chen2024deblurgs}, and COLMAP-Free\citep{fu2024colmapfree3dgs}. Our proposed method effectively reconstructs sharp scenes from blurry inputs, producing fewer artifacts and capturing finer details. Zoom in for a clearer view of the comparison.}
    \label{fig:generation}
\end{figure}
\begin{figure}[ht]
    \centering
    \includegraphics[width=1\linewidth]{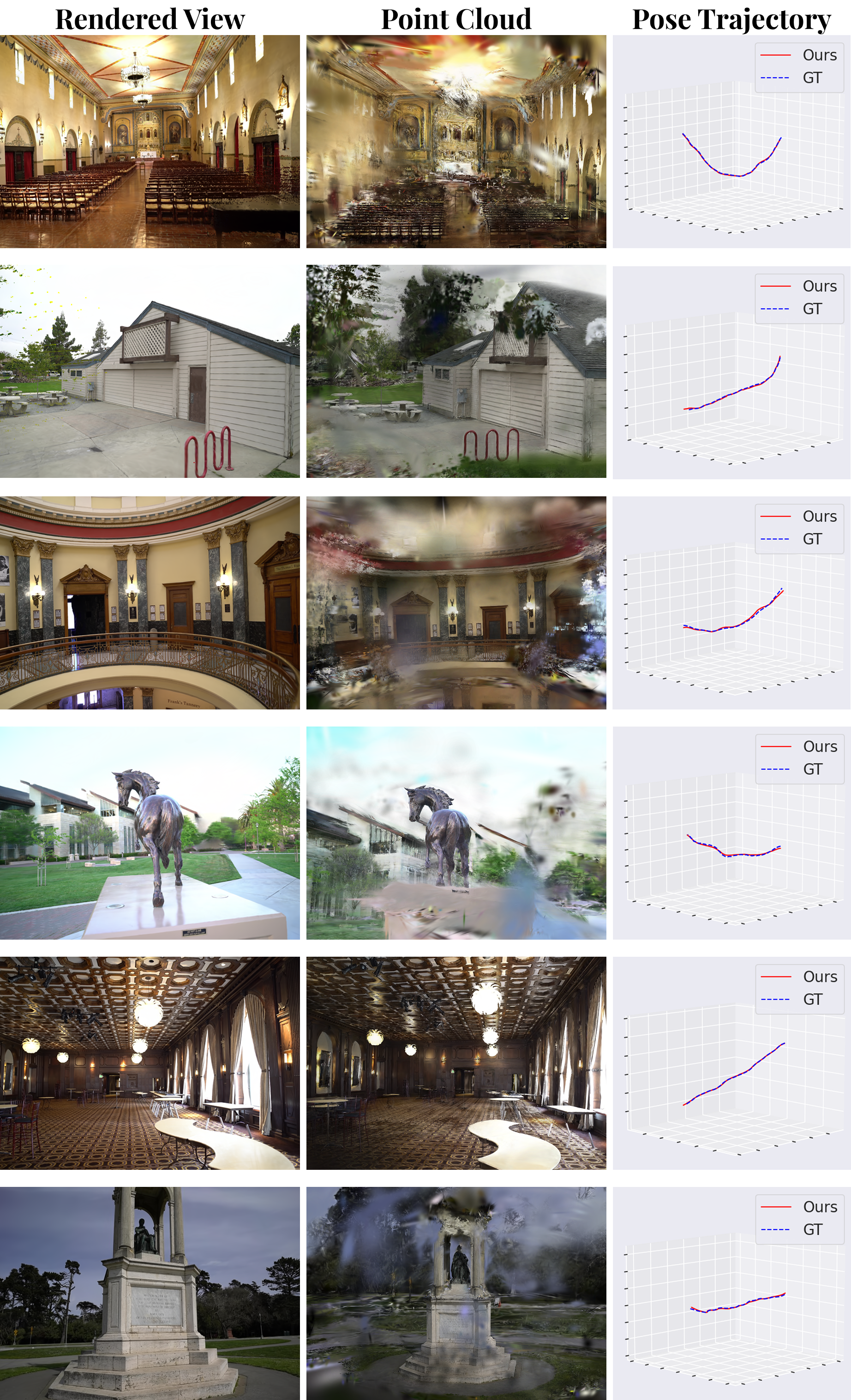}
    \caption{Camera motion estimation results on the Tanks and Temples dataset. For each scene, we present the rendering result (left column), 3D structure from point clouds (middle column), and camera pose estimations (right column).
    Notably, the ExBluRF dataset is specifically tailored for deblurring tasks that involve dynamic scene reconstruction with significant object motion. However, due to its minimal inherent camera motion, camera pose trajectory assessment is excluded from the evaluation.
    }
    \label{fig:cam-pose}
\end{figure}

\section{Qualitative Results} \label{supp:results}
We provide quantitative results on dynamic view synthesis and camera motion estimation in \cref{fig:generation,fig:cam-pose}, respectively.
% Our model captures significant structural and textural details that closely resemble actual physical spaces in \cref{fig:generation}
% While slightly noisier, the additional dense 3D Gaussians preserve sufficient geometric details to reconstruct the scene, as shown in the middle column, enabling accurate synthesis of the appearance and layout of the rendered view in the leftmost column.
% The close alignment between our estimation and the ground truth in the rightmost column indicates accurate motion estimation across scenes, underscoring its effectiveness in challenging environments.
% less informative for quantitative evaluation in this particular experimental configuration.

\section{Ablation Studies} \label{supp:ablation}
We conduct two ablation studies to verify the effectiveness and efficiency of our proposed components.
\subsection{Loss Functions}
% We conduct an ablation study to evaluate the impact of each loss component on dynamic view synthesis by isolating each term.
To evaluate the contribution of each loss component, we conduct an ablation by systematically isolating each loss term and assessing its impact.
% on dynamic view synthesis.
Results in~\cref{tab:loss-functions} show that $\mathcal{L}_\text{image}$ alone is insufficient for dynamic view synthesis from blurry inputs, with pose loss having a slightly greater impact on performance than depth loss.
Depth loss is critical in NeRF-based methods~\citep{bian2023nope}, though they hold less importance in Gaussian-based methods.
% Depth-related losses are critical in advanced pose-unknown approaches, particularly

\subsection{Additional Gaussian Points vs. Training Time}
\begin{figure}[t]
    \centering
    \begin{subfigure}{\columnwidth}
        \centering
        \includegraphics[width=0.49\linewidth]{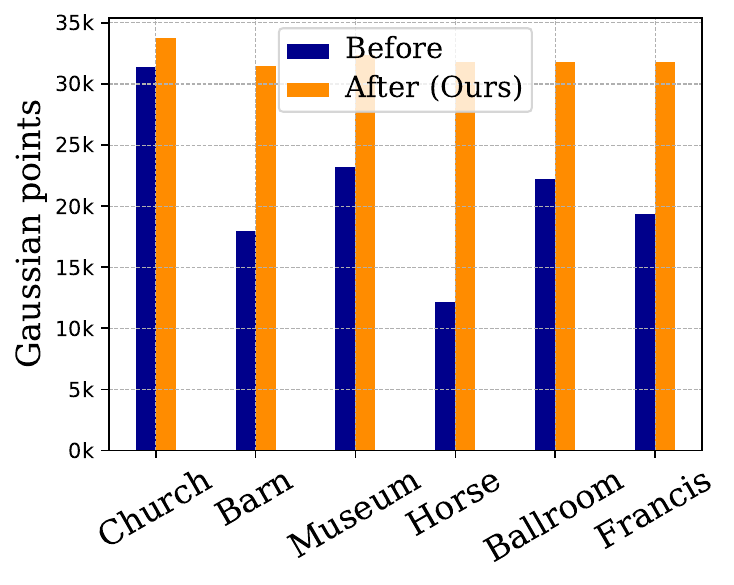}
        \hfill
        \includegraphics[width=0.49\linewidth]{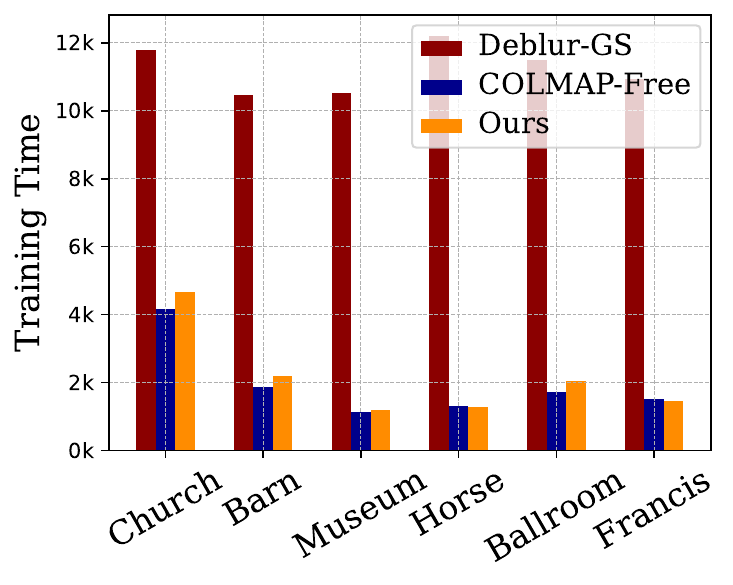}
        % \caption{Tanks and Temples}
        \label{fig:T-gaussians}
    \end{subfigure}
    % \begin{subfigure}{\columnwidth}
    %     \centering
    %     \includegraphics[width=0.49\linewidth]{sec/figure/point_addition_tank.pdf}
    %     \hfill
    %     \includegraphics[width=0.49\linewidth]{sec/figure/training_time_tank.pdf}
    %     \caption{Deblur-NeRF}
    %     \label{fig:D-gaussians}
    % \end{subfigure}
    \caption{Ablation on the number of Gaussian points after converting from sparse to dense Gaussian points and its impact on training time on Tanks and Temples dataset. Training time is converted from \textit{hrs:min:sec} into a decimal format.}
    \label{fig:point-and-time}
\end{figure}

~\cref{fig:point-and-time} illustrates the effect of adding dense Gaussians to the sparse Gaussians on the training time. ``Before'' refers to the sparse Gaussian points initialized from the original 3DGS, while ``After'' represents the dense Gaussians generated through our module.
% We compare the training time of our model with that of two Gaussian-based baseline models.
% Generating additional points requires a slight training time but on average less than an hour.
Our approach is relatively similar to COLMAP-Free~\citep{fu2024colmapfree3dgs}, exhibiting only a marginal difference in training time while generating dense Gaussian points.

\begin{table}[t]
    \centering
    \setlength{\tabcolsep}{1.2mm}
    \resizebox{1\columnwidth}{!}{%
    \begin{tabular}{lcccccc}
    \toprule[1.5pt]
        \multicolumn{1}{c}{} & 
        \multicolumn{3}{c}{Church (Indoor)} &
        \multicolumn{3}{c}{Horse (Outdoor)} \\
        \cmidrule(rl){2-4}
        \cmidrule(rl){5-7}
        \textbf{Model} & {PSNR↑} & {SSIM↑} & {LPIPS↓} & {PSNR↑} & {SSIM↑} & {LPIPS↓}\\
        \midrule[1pt]
         \textbf{Ours} &\textbf{34.04}&\textbf{0.98}&\textbf{0.02}&\textbf{36.52}&\textbf{0.98}&\textbf{0.01}\\
         \midrule
        \quad $\mathcal{L}_\text{image}$ only &31.10&0.93&0.09&28.57&0.97&0.05 \\
        % \quad w/o $\mathcal{L}_\text{depth}$  &33.82&0.85&0.05&34.27&0.98&0.03 \\
        \quad w/o $\mathcal{L}_\text{pose}$   &32.99&0.94&0.08&29.40&0.96&0.04 \\
    \bottomrule[1.5pt]
    \end{tabular}}
    \caption{Ablation study of each loss component on an indoor and outdoor scene from Tanks and Temples. ``Ours'' comprises all the loss terms, and ``w/o $\mathcal{L}_\text{pose}$" utilizes photometric and depth losses.}
    \label{tab:loss-functions}
\end{table}
% \begin{figure}[t]
%     \centering
%     \begin{subfigure}{.49\columnwidth}
%         \centering
%         \includegraphics[width=1\columnwidth]{sec/figure/point_addition_Tank.pdf}
%         \caption{Tank and Temples}
%     \end{subfigure}
%     \begin{subfigure}{.49\columnwidth}
%         \centering
%         \includegraphics[width=1\columnwidth]{sec/figure/point_addition_Tank.pdf}
%         \caption{Deblur-NeRF}
%     \end{subfigure}
%     \caption{Ablation on training time on two datasets. The Y-axis is The time converted from \textit{hrs:min:sec} to a decimal format.}
%     \label{fig:training-time}
% \end{figure}

\section{Future Work}
% In the future, we will advance our focus on adapting our model for real-time applications using smartphone cameras and vision sensors.Additionally, we aim to explore camera pose estimation and motion modeling in real-world scenarios, including robotic navigation and real-time video capture.
% Our po is mostly effective for static scenes and lacks support for dynamic scenes.
In future work, we aim to enhance our approach to novel view synthesis by improving its adaptability and practical applicability, specifically extending 3DGS to dynamic scenes through advanced pose optimization.
These improvements are expected to further enhance rendering quality and robustness, which includes optimizing computational efficiency to enable real-time performance, particularly in smartphone cameras, facilitating seamless and dynamic scene reconstruction in everyday scenarios.

\end{document}